\UseRawInputEncoding
\documentclass[letterpaper]{article} 
\usepackage{2026}  
\usepackage{times}  
\usepackage{helvet}  
\usepackage{courier}  
\usepackage[hyphens]{url}  
\usepackage{graphicx} 
\usepackage{natbib}  
\usepackage{caption} 
\usepackage{algorithm}
\usepackage{algorithmic}

\usepackage{newfloat}
\usepackage{listings}
\DeclareCaptionStyle{ruled}{labelfont=normalfont,labelsep=colon,strut=off} 
\usepackage{newfloat}
\usepackage{listings}
\usepackage{mathrsfs}
\usepackage{amsmath}
\usepackage{amssymb}
\usepackage{float}
\usepackage{booktabs}
\usepackage{multirow} 
\floatstyle{ruled}
\newfloat{listing}{tb}{lst}{}
\floatname{listing}{Listing}

\title{Anti-Inpainting: A Proactive Defense Approach against Malicious Diffusion-based Inpainters under Unknown Conditions}
\author {
    Yimao Guo\textsuperscript{\rm 1},
    Zuomin Qu\textsuperscript{\rm 1},
    Wei Lu \footnotemark[1] \textsuperscript{\rm 1},
    Xiangyang Luo \thanks{Corresponding authors} \textsuperscript{\rm 2}
}
\affiliations {
    \textsuperscript{\rm 1}Sun Yat-sen University, Guangzhou, China\\
    \textsuperscript{\rm 2}State Key Laboratory of Mathematical Engineering and Advanced Computing, Zhengzhou, China\\
    guoym39@mail2.sysu.edu.cn, 
    quzm@mail2.sysu.edu.cn, 
    luwei3@mail.sysu.edu.cn, 
    luoxy\_ieu@sina.com
}

\usepackage{bibentry}

\begin{document}

\maketitle

\begin{abstract}
	With the increasing prevalence of diffusion-based malicious image manipulation, existing proactive defense methods struggle to safeguard images against tampering under unknown conditions. 
	To address this, we propose Anti-Inpainting, a proactive defense approach that achieves protection comprising three novel modules. 
	First, we introduce a multi-level deep feature extractor to obtain intricate features from the diffusion denoising process, enhancing protective effectiveness. Second, we design a multi-scale, semantic-preserving data augmentation technique to enhance the transferability of adversarial perturbations across unknown conditions. Finally, we propose a selection-based distribution deviation optimization strategy to bolster protection against manipulations guided by diverse random seeds. Extensive experiments on InpaintGuardBench and CelebA-HQ demonstrate that Anti-Inpainting effectively defends against diffusion-based inpainters under unknown conditions. Additionally, our approach demonstrates robustness against various image purification methods and transferability across different diffusion model versions.
\end{abstract}
\section{Introduction}
Recent advancements in diffusion models have enabled remarkable progress in high-fidelity content generation, making the distinction between synthetic and authentic content increasingly difficult \cite{couairon2023diffedit, meng2022sdedit}. Specifically, the latent diffusion model (LDM) excels at controllable image manipulation \cite{Rombach_2022_CVPR}. LDM's efficiency stems from its operation within a compressed latent space, where a U-Net architecture performs iterative denoising \cite{10.1007/978-3-319-24574-4_28}. Moreover, diffusion-based inpainting techniques empower users to specify manipulation regions via masks, yielding highly authentic results through fine-grained control \cite{XIANG2023109046}.
\begin{figure}[!t]
	\includegraphics[width=\linewidth]{ 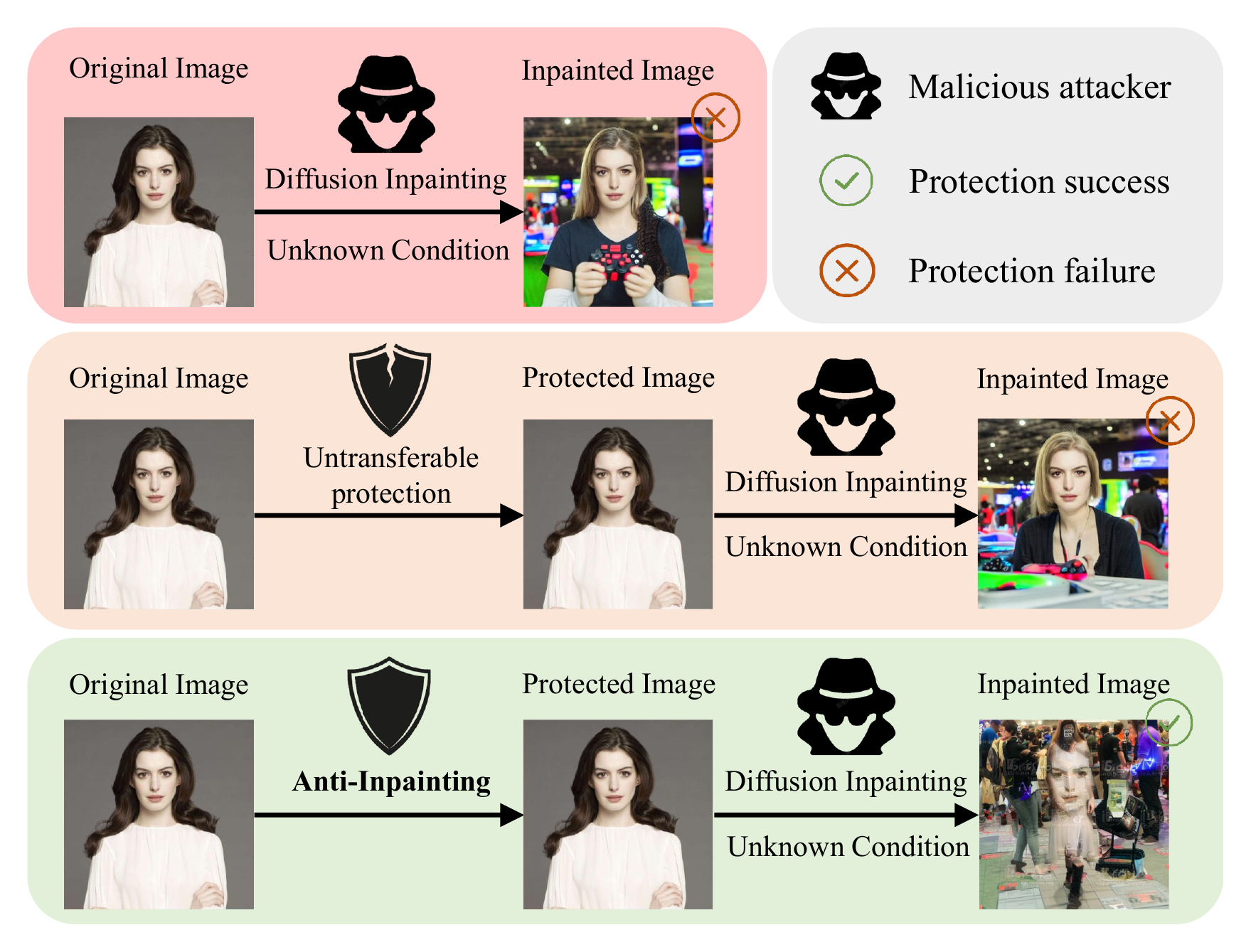}
	\caption{The proactive defense against the misuse of diffusion models guided by unknown conditions.}
	\label{fig1}
\end{figure}

However, these advancements also introduce significant ethical concerns regarding the malicious use of diffusion-based image manipulation \cite{10.1007/978-3-031-73036-8_8}. Capable of producing hyper-realistic and persuasive outputs, image manipulation models \cite{wang2023imagen} can be exploited to fabricate news, disseminate disinformation, and craft misleading imagery, as shown in Figure \ref{fig1} (top row). For instance, open-source diffusion models \cite{Brooks_2023_CVPR} allow for the effortless fabrication of scenarios, such as the false portrayal of a celebrity's arrest. Therefore, as these models grow in sophistication, developing robust safeguards against such misuse becomes imperative.

Proactive defense methods \cite{wang2025diffusion, pmlr-v202-liang23g, phan2025latent, mi2024visual} using adversarial perturbations have recently emerged as a promising strategy to counter the misuse of diffusion models. 
However, a critical flaw in most current methods is their failure to address \textbf{unknown conditions}—scenarios where an attacker can specify arbitrary manipulation regions and iterate through different initial latent states, as shown in Figure \ref{fig1} (middle row). 
Although some work has begun to tackle the challenge of unknown masks via augmentation, they have not fully addressed the threat of latent state resampling. 
This vulnerability can be exploited by attackers to bypass existing defenses and generate high-quality unauthorized manipulations \cite{hertz2023prompttoprompt, Zhang_2023_ICCV}.

To address these challenges, this paper presents Anti-Inpainting, a proactive defense approach designed to protect images from diffusion-based inpainting under unknown conditions, as depicted in Figure \ref{fig1} (bottom row). Our method introduces three key innovations.
Firstly, we enhance the perturbation's effectiveness by shifting the adversarial target. In the diffusion process, the U-Net module predicts noise by attending to multi-level features of the input. We identified that features more critical to the manipulation process exhibit larger gradients with respect to the predicted noise. Therefore, instead of attacking the final predicted noise, we directly target these crucial multi-level deep features. 
Furthermore, to counter manipulations under unknown masks, we integrate multi-scale, semantic-preserving data augmentation into the optimization process, thereby improving the perturbation's robustness.
Finally, we mitigate the impact of latent state randomness. The initial latent state, a random variable, is a key input to the U-Net that significantly influences its noise prediction. To address this, we propose a selection-based distribution deviation optimization strategy. This strategy identifies latent states that are prone to causing protection failures and specifically focuses the optimization on them. By doing so, we reduce the impact of randomness and enhance the consistent protective performance of our adversarial samples.
We summarize our main contributions as follows:
\begin{itemize}
	\item We propose Anti-Inpainting, a proactive defense approach that generates adversarial perturbations to protect images against diffusion-based inpainting models under unknown conditions.
	\item We introduce a multi-level feature extractor to capture hierarchical image features. To enhance the transferability of perturbations, we design a multi-scale, semantic-preserving data augmentation. Furthermore, we develop a selection-based distribution deviation optimization strategy to ensure both effective protection and efficient optimization.
	\item Extensive experiments demonstrate that our proposed Anti-Inpainting effectively safeguards images against various diffusion-based inpainting models and exhibits strong robustness to diverse image purification techniques.
\end{itemize}

\section{Related Work}
\subsection{Diffusion Model}
Diffusion models have rapidly become a cornerstone of modern generative AI, led by the paradigm of Denoising Diffusion Probabilistic Models (DDPMs) \cite{NEURIPS2020_4c5bcfec}. These models learn to synthesize data by reversing a gradual noising process \cite{Bansal_2023_CVPR, 10.1145/3592116, gal2023an}. A pivotal advancement was the introduction of LDMs, which apply the diffusion process in a compressed latent space, drastically improving computational efficiency and enabling high-fidelity synthesis \cite{ Li_2023_CVPR, Ruiz_2023_CVPR}. 
Moreover, techniques like inpainting mask guidance have provided robust control over the generation process, making image manipulation powerful and widespread.

\subsection{Proactive Defense Model}
Recent studies have introduced adversarial perturbations to protect images from unauthorized edits by diffusion-based models \cite{liang2023mist, wang2024simac, xue2023toward, xu2024perturbing, jeong2025faceshielddefendingfacialimage, lo2024distraction, van2023anti}. A key method, PhotoGuard \cite{pmlr-v202-salman23a}, disrupts the generative process through dual attacks on the model's encoder and diffusion stages via latent space manipulation. However, its effectiveness is largely confined to known attack conditions (e.g., predefined masks) and falters against unforeseen manipulations, such as those involving manually created masks.
To enhance protection against varied mask shapes, DiffusionGuard \cite{ choi2025diffusionguard} introduces contour-shrinking mask augmentation. Despite these advances, a broader limitation of existing methods is their lack of attention to other crucial conditions in the generation process, such as the initial latent state.

\section{Preliminaries}
\paragraph{Threat Model} 
Image inpainting, which involves modifying targeted regions within an image, is another significant application of generative models. The process begins by applying a mask $M$ to the manipulation region of a given image $I$. An image encoder ~$ \mathcal{E}(\cdot) $ is then used to extract embeddings from $I$. In the subsequent diffusion process, these embeddings and the mask are concatenated with the latent state $z_t$, serving as input to the noise predictor ~$\epsilon_\theta(\cdot)$. This iterative denoising process for inpainting can be formulated as follows:
\begin{equation}
	z_{t-1} = \frac{1}{\sqrt{\alpha_t}}\left( z_t - \frac{1-\alpha_t}{\sqrt{1-\bar\alpha_t}} {n_{pred}} \right) + \sigma_t {n}
	\label{eq3},
\end{equation}
\begin{equation}
	{n_{pred}} = \epsilon_\theta(z_t, \mathcal{E}(I), M, t, clip(\mathscr{T}))
	\label{eq3-2}.
\end{equation}
where~$ t $~denotes the timestep, ~$\alpha_t$~ and ~$ \bar\alpha_t $~are pre-defined hyper-parameters, and ~$clip(\mathscr{T})$~ represents the text embeddings for the manipulation prompt ~$ \mathscr{T} $~.

\paragraph{Task Formulation} 
The goal of proactive defense is to protect image privacy by disrupting unauthorized manipulations performed by diffusion models.
Given a clean image  $I$ and a diffusion model, the adversarial perturbation $\delta$ is added into the clean image $I$. 
To ensure visual imperceptibility, the perturbation $\delta $ is typically limited using the norm bound~$ \eta$. 
To disrupt the reverse diffusion process, the perturbation ~$ \delta $~ is optimized by:
\begin{equation}
	\delta = \mathop{\arg\max}\limits_{||\delta||_{\infty}\leq \eta}~ 
	\left\| n_{pred} - \epsilon_{\theta}(z, \mathcal{E}(I + \delta), M, t, clip(\mathscr{T})) \right\|_2
	\label{eq5}.
\end{equation}

\begin{figure*}[!t]
	\centering
	\includegraphics[width= 0.97\linewidth]{ 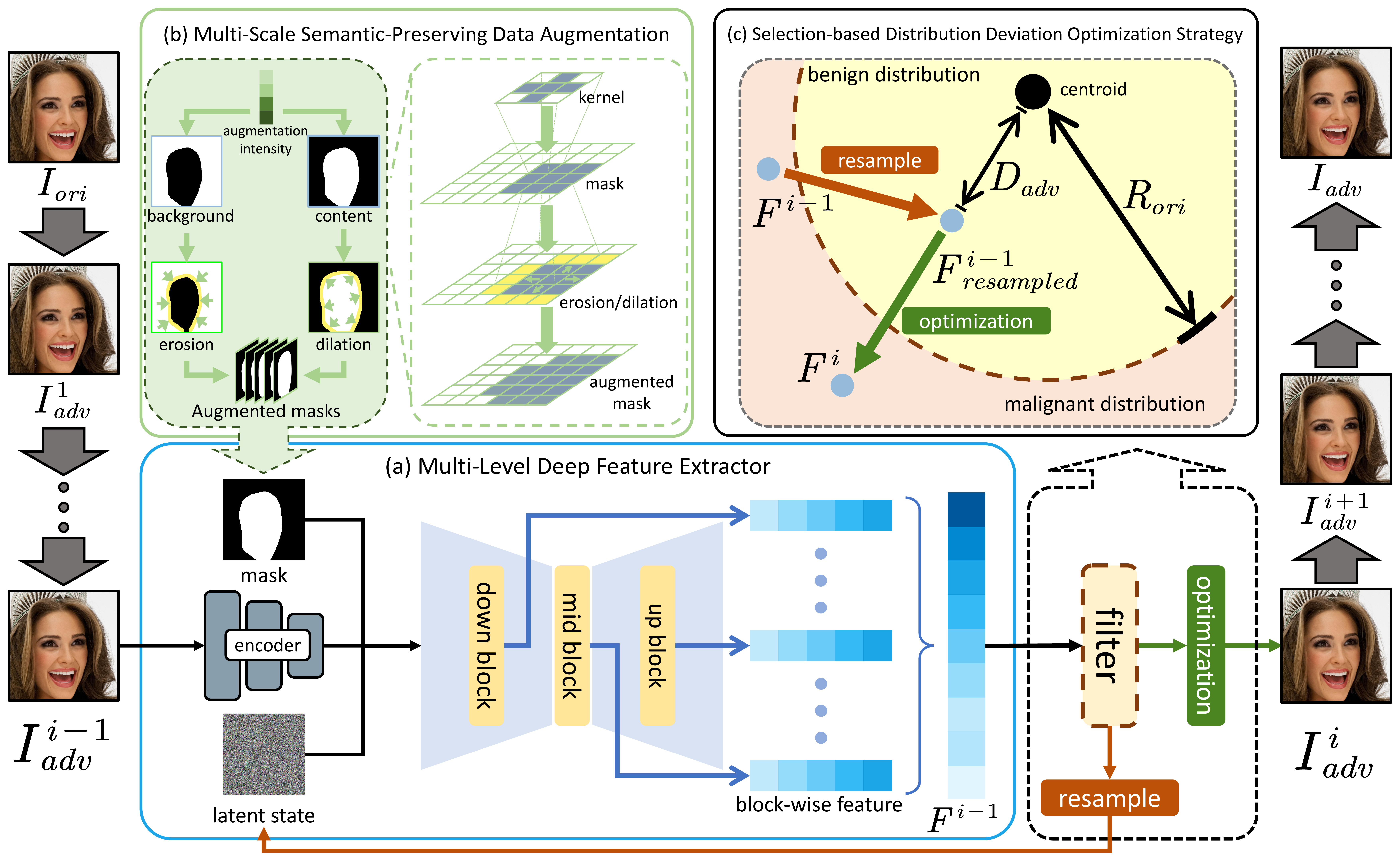}
	\caption{
		Overview of Anti-Inpainting. 
		We propose an iterative method to generate adversarial images from  original images. In each iteration, diverse masks are first generated via multi-scale semantic-preserving data augmentation. These masks, along with the latent state and the adversarial image, are fed into a multi-level deep feature extractor. A selection-based distribution deviation optimization strategy then selects salient features from the extractor, which are subsequently used to update the adversarial image.
	}
	\label{fig2}
\end{figure*}
\section{Method}
In this section, we introduce Anti-Inpainting, a proactive defense method designed to safeguard images against manipulation by inpainting models. Our approach integrates three key components: a multi-level deep feature extractor, multi-scale semantic-preserving data augmentation, and a selection-based distribution deviation optimization strategy. The overall workflow of our algorithm is illustrated in Figure \ref{fig2}.
Our method builds upon Projected Gradient Descent (PGD) framework \cite{Madry:2017tvh}, an iterative adversarial attack method. Each iteration of our approach begins with the multi-scale semantic-preserving data augmentation, which provides diverse masks to the U-Net within the diffusion module. Subsequently, the multi-level deep feature extractor extracts features from the U-Net as it processes the augmented data. 
Finally, the selection-based distribution deviation optimization strategy selects specific features from the extractor and computes the loss function to update the adversarial perturbation.


\subsection{Multi-Level Deep Feature Extractor}
\label{Multi-Level Deep Feature Extractor}
Mainstream adversarial attacks on diffusion models primarily target the final output of the U-Net, the predicted noise. This approach implicitly assumes the final prediction is a sufficient proxy for all crucial internal computations. We contend that the internal feature maps of the U-Net's encoder and decoder blocks offer a more comprehensive target. These maps represent a spectrum of features, from low-level patterns to high-level semantics, which are vital for the denoising process. By only attacking the final output, existing methods fail to fully exploit vulnerabilities within the model's feature hierarchy. Therefore, to achieve a more potent attack, we propose a multi-level deep feature extractor that captures block-wise features across the U-Net architecture, as shown in Figure \ref{fig2} (a).

Firstly, we construct the inputs for  the first block of the U-Net module, denoted as $\epsilon^{0}$. At each denoising timestep $t$, the model takes two primary inputs: a main input tensor and a conditioning vector $c$. The main input tensor is formed by concatenating the initial latent state $z$, the VAE-encoded input image $\mathcal{E}(I_{input})$, and the input mask $M_{input}$. The conditioning vector $c$ combines the timestep embedding for $t$ and the text prompt embedding, ${clip}(\mathscr{T})$:
\begin{equation}
	{c} =  {concat}(t,  {clip} (\mathscr{T})),
	\label{eq6}
\end{equation}
\begin{equation}
	f_{0} = \epsilon^{0}(z, \mathcal{E}(I_{input}), M_{input}, c),
	\label{eq7}
\end{equation}
where the latent state $ z $ is sampled from normal distribution, and timestep $t$ is sampled from uniform distribution. We obtain the intermediate variable from the U-Net module, as follows:
\begin{equation}
	f_{i} = 
	\left
	\{\begin{array}{lll}
		\epsilon_{down}^{i}\left(f_{i-1}, {c}\right), \quad   & if & 5 \textgreater i \textgreater 0, \\
		\epsilon_{mid}^{i}\left(f_{i-1}, {c}\right),\quad  ~& if & i=5, \\
		\epsilon_{up}^{i}(f_{i-1}, f_{10-i}, {c}),\quad ~& if & 10 \textgreater i \textgreater 5,
	\end{array}
	\right.
	\label{eq8}
\end{equation}
where $i$ is the number corresponding to the block $\epsilon$.
$\epsilon_{down}^{i}$, $\epsilon_{mid}^{i}$ and $\epsilon_{up}^{i}$ denote the downsampling block, middle layer, and upsampling block of U-Net, respectively.
And then $f_{10}$ is the output of post-processing module $\epsilon^{10}$. In addition, we combine the intermediate variable of each U-Net block and define the whole process as multi-level deep feature extraction $\phi$, which is defined as:
\begin{equation}
	\begin{aligned}
		F &= \phi \left(z, \mathcal{E}(I_{input}), M_{input}, t, clip(\mathscr{T})\right),  \\
		&= concat\left(f_{0},f_{1}, f_{2}, f_{3}, \ldots \ldots f_{10}\right).
	\end{aligned}
	\label{eq9}
\end{equation}

We compute the mean of the multi-level deep features,  $ \overline{F_{{ori}}} $, across multiple latent states $z$ to serve as the feature representation for the original image $I_{ori}$. Additionally, we compute the distribution radius $R_{ori}$ to quantify the dispersion of these features for the image $I_{ori}$:
\begin{equation}
	\left.\overline{F_{{ori}}}=E_{z \in N(0,1)} [ \phi \left(z, \mathcal{E}(I_{ori}), M , t, \mathscr{T}\right)\right],
	\label{eq10}
\end{equation}
\begin{equation}
	\left.R_{ori}=E_{z \in N(0,1)} [ \| \phi \left(z, \mathcal{E}(I_{ori}),  M, t, \mathscr{T} \right)-\overline{F_{{ori}}} \|_{2}\right].
	\label{eq11}
\end{equation}
\subsection{Multi-Scale Semantic-Preserving Data Augmentation}
\label{Multi-Scale Semantic-Preserving Data Augmentation}
A key limitation of current methods is their reliance on known guidance conditions, leaving images vulnerable to the unpredictable and multifaceted manipulations employed by malicious users. While DiffusionGuard \cite{ choi2025diffusionguard} enhances robustness against unknown conditions by using augmented masks, its contour-shrinking technique compromises the mask's semantic integrity, thus weakening the overall protection. To overcome this, we introduce multi-scale semantic-preserving data augmentation, as shown in Figure \ref{fig2} (b). Our method enhances the diversity of masks used in adversarial optimization without sacrificing their semantic information:
\begin{equation}
	M_{aug}=
	\left\{
	\begin{array}{lll}
		\Omega \ ( \ M , n) \ , \quad   &if& n \geq 0, \\
		
		\zeta \  ( \ M , -n) \ , \quad &if& n \textless 0,
	\end{array}
	\right.
	\label{eq12}
\end{equation}
where ${\Omega}(\cdot)$ is mask dilation operation, and ${\zeta}(\cdot)$ is mask erosion operation\footnotemark. 
We introduce a data augmentation scheme for the input mask $M$. The process is governed by an integer $n$ sampled uniformly from $[-\gamma, \gamma]$, where $\gamma$ is the augmentation intensity hyperparameter. The augmented mask $M_{aug}$ is obtained by applying either a morphological dilation (for $n \textgreater 0$) or erosion (for $n \textless 0$) to $M$ using a square kernel of size $|n| \times |n|$. 
By applying the moderate dilation or erosion, we maintain the mask's overall topology and primary shape, ensuring it continues to represent the same semantic object while introducing boundary variations for adversarial optimization.
\footnotetext{https://docs.opencv.org/}
\subsection{Selection-based Distribution Deviation Optimization Strategy}
\label{Selection-Based Distribution Deviation Optimization Strategy}

To maximize the protective effect under a fixed perturbation budget $\eta$, we address the issue of inefficient budget allocation. We posit that optimizing against adversarial features that have already deviated drastically from the original distribution yields diminishing returns. Therefore, we introduce a selection mechanism to focus the optimization on more impactful features.
Specifically, instead of indiscriminately optimizing against all adversarial features, we focus the optimization process only on the adversarial features that remain within a defined proximity of the benign feature distribution. By avoiding budget allocation to features that have already deviated significantly, we can achieve a more potent and robust protective effect.


To implement this strategy, we first characterize the benign feature space. As illustrated in Figure \ref{fig2} (c), we take the original clean image and generate a set of benign variants. These variants are then passed through the deep feature extractor $\phi$ to obtain a collection of benign features. From this collection, we compute the centroid $\overline{F_{{ori}}}$ and a boundary threshold $R_{ori}$, which together define the boundary of our target benign distribution.
In each optimization step $i$, this benign distribution is used as a reference. We take the adversarial image from the previous iteration, $I_{adv}^{i-1}$, and extract its corresponding adversarial feature using the same extractor $\phi$: 
\begin{equation}
	F^{i-1}=\phi \left(z, \mathcal{E}(I_{adv}^{i-1}), M_{aug} , t, \mathscr{T}\right) ,
	\label{eq13}
\end{equation}
where $z$ is the latent state. 
Our selection mechanism operates as a conditional filter within each optimization iteration $i$. For each adversarial feature sample, generated using a latent state $z$, we first determine its viability for optimization.
This is done by comparing its distance from the benign centroid, $D_{adv} = ||F^i - \overline{F_{ori}}||_2$, against a dynamic threshold, $\tau \cdot R_{ori}$, where $\tau$ is a hyperparameter.
If the feature is outside the boundary ($D_{adv} \textgreater \tau \cdot R_{ori}$), as exemplified by the red point in Figure \ref{fig2}, we consider it an inefficient candidate for optimization. We discard this sample and resample a new latent state, $z$, to generate a new feature. This process repeats until a viable candidate is found or a maximum number of resampling attempts is reached. This prevents wasting the perturbation budget on features that have already diverged excessively.
Conversely, if a feature lies within the boundary, it is deemed eligible for the adversarial attack. The goal of the attack is to push this feature away from the benign distribution. To achieve this, we define our loss function to maximize the distance between the adversarial feature $F^{i}$ and the benign centroid $\overline{F_{ori}}$, as shown below:
\begin{equation}
	L_{adv} =-\left\|\left(F^{i}-\overline{F_{ori}}\right) \right\|_{2}^{2}.
	\label{eq15}
\end{equation}
This loss guides the update of the adversarial image. In each step, the gradient of $L_{adv}$ with respect to the adversarial image is computed and used to perform the update.

\begin{figure*}[!t]
	\centering
	\includegraphics[width= \linewidth]{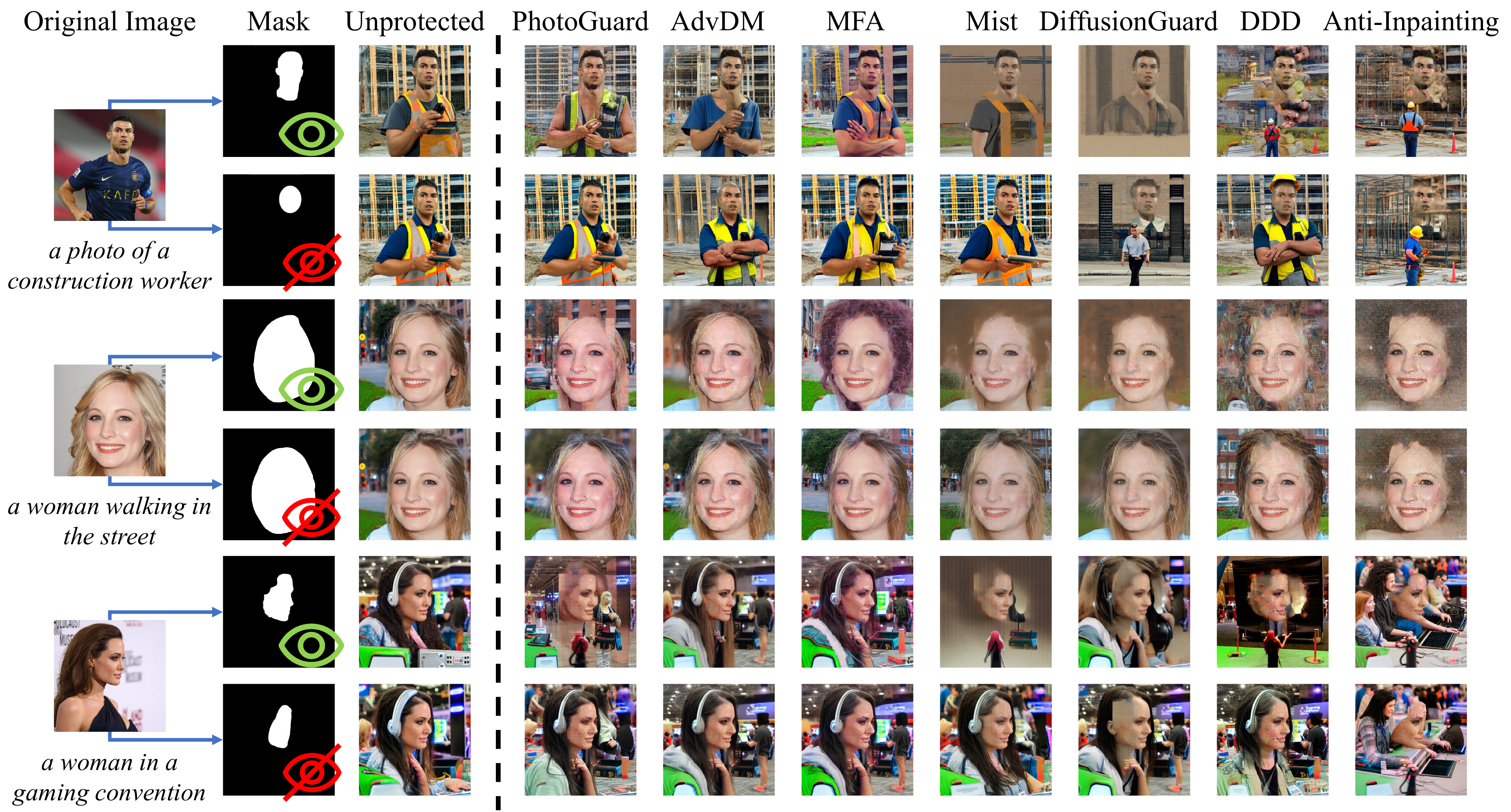}
	\caption{The qualitative results between comparison methods and Anti-Inpainting. The text below each original image is the prompt used to generate the corresponding forged image. The green eye icon on the mask indicates that the mask is known during adversarial example generation, while the red, crossed-out eye icon signifies that it is unknown.}
	\label{fig3}
\end{figure*}
\section{Experiments}
\subsection{Experimental Setup}
\paragraph{Datasets } We conduct quantitative evaluations of our approach and competing methods on the InpaintGuardBench \cite{ choi2025diffusionguard} and CelebA-HQ \cite{karras2018progressive} datasets. InpaintGuardBench consists of 42 images, each containing one known mask, four unknown masks, and 10 text prompts. For CelebA-HQ, we select the first 100 images for testing. For each of these images, we use the corresponding skin mask from CelebAMask-HQ \cite{Lee_2020_CVPR} as the known mask and manually generated four unknown masks. These masks are created manually by either drawing with a circular brush or overlaying simple geometric shapes. Finally, all masks used in the quantitative experiments will be made publicly available on our project repository.

\paragraph{Comparison Methods } We compare six adversarial proactive defense methods for diffusion models: PhotoGuard, AdvDM, MFA \cite{Yu_Chen_Ding_Zhang_Tang_Ma_2024}, Mist, DiffusionGuard, and DDD \cite{10.24963/ijcai.2024/856}. 
To simulate real-world scenarios, we generate the protected images using skin masks and empty text prompts. Subsequently, we evaluate the protective performance of these images against inpainting attacks guided by manual masks and malicious text prompts.

\paragraph{Evaluation Metrics } We assess the impact of adversarial perturbations on diffusion-based inpainting models using three sets of metrics. To quantify the difference between the protected and unprotected inpainting results under the same random seed, we compute PSNR, SSIM \cite{1284395}, and LPIPS \cite{Zhang_2018_CVPR}. The visual quality of the resulting images is evaluated via the ImageReward (IR) score \cite{NEURIPS2023_33646ef0}. Lastly, the ArcFace similarity (ARC) \cite{Deng_2019_CVPR} is calculated to evaluate the preservation of facial identity information.

\paragraph{Implementation Details } 
The perturbation is constrained under the L-infinity norm with a magnitude of 16/255. For each sample, we perform 800 optimization iterations. Our primary attack target is the Runway v1.5 diffusion-based inpainter. To evaluate transferability, we also test the generated adversarial samples on the Stability AI v2.0 inpainter \cite{Rombach_2022_CVPR}. All experiments are conducted on NVIDIA 3090 GPUs, and our approach requires up to 16GB of GPU memory.

\subsection{Comparative Experiment}
\paragraph{Qualitative Results} Figure \ref{fig3} presents the original images, their corresponding masks, and the resulting inpainted images. The figure also displays the inpainting results from adversarial examples generated by both the compared methods and our proposed approach. As shown, the compared methods are effective under known conditions but fail under unknown ones. In contrast, our approach demonstrates strong protective performance in both scenarios. These qualitative results validate our conclusion that increasing mask diversity during adversarial training and strategically selecting the initial latent state improves the transferability of adversarial examples to unknown conditions.
\begin{table*}[!t]
	\centering
	\small
	\label{table1}
	\begin{tabular}{lcccccccccc}
		\toprule
		\multirow{2}{*}{{Methods}} & 
		\multicolumn{5}{c}{{InpaintGuardBench}} & 
		\multicolumn{5}{c}{{CelebA-HQ}} \\
		\cmidrule(l){2-6} \cmidrule(l){7-11}
		& {PSNR$\downarrow$} & {SSIM$\downarrow$} & {LPIPS$\uparrow$} & {IR$\downarrow$} & {ARC $\downarrow$} 
		& {PSNR$\downarrow$} & {SSIM$\downarrow$} & {LPIPS$\uparrow$} & {IR$\downarrow$} & {ARC $\downarrow$} \\
		\midrule
		PhotoGuard  
		& 16.518 & 0.600 & 0.404 & -0.015 & 0.674
		& 17.874 & 0.640 & 0.380 & -0.011 & 0.861 \\
		AdvDM       
		& 16.695 & 0.598 & 0.402 & -0.032 & 0.677
		& 18.000 & 0.600 & 0.393 & -0.011 & 0.836 \\
		MFA         
		& 16.975 & 0.609 & 0.392 & -0.032 & 0.739
		& 19.372 & 0.686 & 0.285 & -0.010 & 0.923 \\
		Mist        
		& 15.687 & 0.533 & 0.481 & -0.274 & 0.635
		& 16.711 & 0.551 & 0.457 & -0.132 & 0.809 \\
		DiffusionGuard 
		& 14.797 & {0.477} & {0.576} & \textbf{-0.578} & 0.571
		& 15.874 & {0.495} & {0.588} & {-1.617} & 0.762 \\
		DDD         
		& {14.390} & 0.488 & 0.520 & -0.224 & 0.548
		& {15.779} & 0.525 & 0.491 & -1.369 & 0.777 \\
		Anti-Inpainting        
		& \textbf{12.875} 
		& \textbf{0.414} 
		& \textbf{0.595} 
		& {-0.473} 
		& \textbf{0.491} 
		& \textbf{14.704} 
		& \textbf{0.468} 
		& \textbf{0.592} 
		& \textbf{-1.658} 
		& \textbf{0.744} \\
		\bottomrule
	\end{tabular}
	\label{table1}
	\caption{The quantitative results of comparison methods and Anti-Inpainting. The best attacking performances of methods are marked as bold.}
\end{table*}
\begin{table*}[!t]
	\centering
	\small
	\label{tab:results}
	\begin{tabular}{lcccccccccc}
		\toprule
		\multirow{2}{*}{{Methods}} & 
		\multicolumn{5}{c}{{InpaintGuardBench}} & 
		\multicolumn{5}{c}{{CelebA-HQ}} \\
		\cmidrule(l){2-6} \cmidrule(l){7-11}
		& {PSNR$\downarrow$} & {SSIM$\downarrow$} & {LPIPS$\uparrow$} & {IR$\downarrow$} & {ARC $\downarrow$} 
		& {PSNR$\downarrow$} & {SSIM$\downarrow$} & {LPIPS$\uparrow$} & {IR$\downarrow$} & {ARC $\downarrow$} \\
		\midrule
		PhotoGuard  
		& 19.964 & 0.722 & 0.297& -0.661 & 0.877 
		& 20.580 &	0.761 &	0.243 &	0.975  & 0.692 
		\\
		AdvDM       
		& 19.751 & 0.687 & 0.305 & -0.717 & 0.853 
		& 20.979 &	0.770 &	0.234 &	0.920 & 0.692 
		\\
		MFA         
		& 22.422 & 0.796 & 0.186 & -0.640 & 0.938 
		& 21.097 &	0.772 &	0.235 &	0.933  & 0.749 
		\\
		Mist        
		& 18.527 & 0.668 & 0.339 & -0.786 & 0.831 
		&19.255 &	0.698 &	0.311 &	0.877  & 0.641 
		\\
		DiffusionGuard 
		& 17.965 & 0.625 & 0.434 & \textbf{-1.076}& \textbf{0.778}
		& 18.241  & 0.647 & 0.386 & 0.729 & 0.604 
		\\ 
		DDD         
		& 21.753 & 0.770 & 0.201 & -0.606 & 0.936 
		&17.509 &	0.652 &	0.378 &	0.881 & 0.567 \\
		Anti-Inpainting        
		& \textbf{17.665} 
		& \textbf{0.618} 
		& \textbf{0.408}
		& -1.012 
		& 0.805 
		& \textbf{15.619} 
		& \textbf{0.573} 
		& \textbf{0.466} 
		& \textbf{0.663} 
		& \textbf{0.476} \\
		\bottomrule
	\end{tabular}
	\caption{The quantitative results of comparison methods and Anti-Inpainting under multiple initial latent states.}
	\label{table2}
\end{table*}

\begin{table}[!t]
	\centering
	\begin{tabular}{lcc}
		\toprule
		{Method} & {runtime(s)} & {GPU memory(MB)} \\ 
		\midrule
		photoguard & 341.43 & 13841  \\ 
		advDM & 252.36  & \textbf{11241}  \\ 
		MFA & 276.31  & 11263  \\ 
		mist & 239.29  & 11871  \\ 
		ddd & 205.26  & 17049  \\ 
		ours & \textbf{180.00}  & 13275  \\ 
		\bottomrule
	\end{tabular}
	\caption{The computational cost of comparison methods and Anti-Inpainting.}
\end{table}

\paragraph{Quantitative Results} 
Table 1 presents the quantitative results of our approach against mainstream methods. Our approach achieves superior performance on PSNR, SSIM, and LPIPS metrics. In terms of the visual quality of the manipulated results (IR), our approach ranks second on InpaintingGuardBench and first on CelebA-HQ. Furthermore, our approach is most effective at disrupting face identity information (ARC) in manipulated images across both datasets.
To simulate robust malicious attacks, we applied tampering five times on InpaintingGuardBench and twenty times on CelebA-HQ, each with a different random seed. We then selected the most successfully tampered outcome for final evaluation. As shown in Table \ref{table2}, leveraging the proposed Selection-based Distribution Deviation Optimization Strategy, our approach obtains the top results across all metrics on CelebA-HQ and secures either the best or second-best performance on all metrics within InpaintingGuardBench.

\paragraph{Computational Cost}
We benchmarked the computational cost on InpaintGuardBench. As detailed in Table 3, our approach, enabled by a selection-based distribution deviation optimization, records the fastest execution time while preserving a GPU memory footprint comparable to that of competing methods.
\begin{figure}[!t]
	\centering
	\includegraphics[width=\linewidth]{ 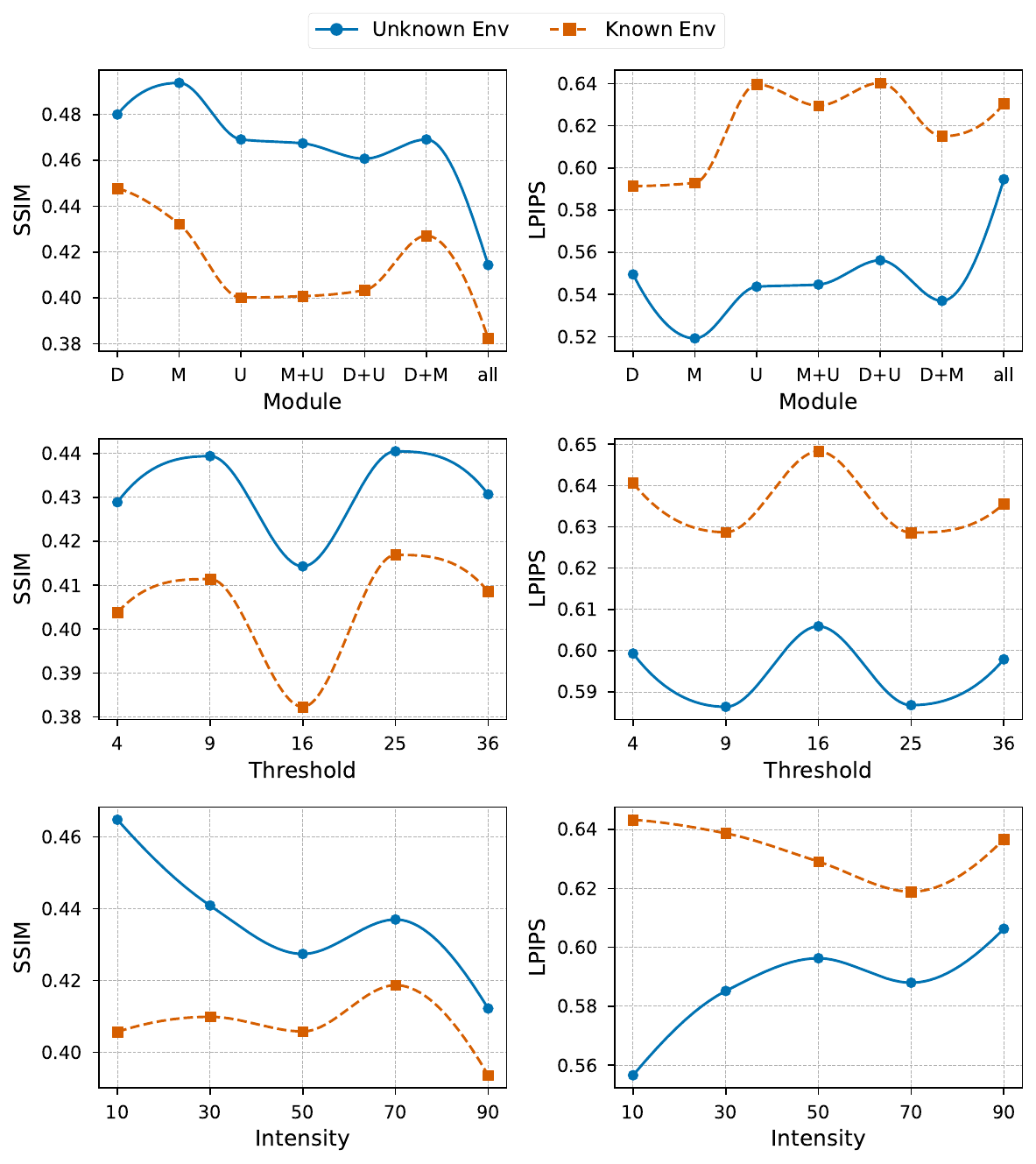}
	\caption{The results of ablation experiments on feature selection, augmentation intensities, and optimization thresholds}
	\label{fig4}
\end{figure}
\subsection{Ablation Study}
\paragraph{Feature Selection}
We conducted an ablation study to evaluate the effectiveness of using multi-level features from the diffusion model for adversarial protection. The U-Net was divided into three components: downsampling blocks $\mathcal{D}$, a middle block~$\mathcal{M}$, and upsampling blocks~$\mathcal{U}$. We then created several control groups by combining features from these respective components. The results reveal the crucial role of features across all U-Net levels.  Specifically, both low-level features from the downsampling path and high-level features from the upsampling path proved essential for generating adversarial examples. This finding underscores the importance of the multi-level feature extractor in our approach.
\paragraph{Optimization Thresholds} 
This ablation study investigates the impact of the optimization threshold on the efficacy of adversarial examples. We hypothesized a U-shaped performance curve: trivially small thresholds would result in futile optimization, while excessively large ones would overlook valuable initial latent states, both diminishing performance. Our results confirm this hypothesis. Significantly, we discovered a strong correlation between the black-box and white-box performance of the adversarial examples across various thresholds in Figure \ref{fig4}. This correlation allows us to use the more readily available white-box metrics as a proxy for tuning the optimization threshold, thereby maximizing the success rate of black-box attacks.
\paragraph{Augmentation Intensities} 
In this ablation study, we evaluated the effect of data augmentation intensity on adversarial example performance.  Our findings indicate that with increasing data augmentation intensity, the white-box performance initially declines before rising. In contrast, the black-box performance exhibits a consistent upward trend. This trend in the white-box setting suggests that our augmentation module does more than simply enhance black-box transferability;  it fundamentally improves the adversarial examples' ability to interfere with the diffusion model's image comprehension.
\subsection{Transferability Study}
\begin{table}[!t]
	\centering
	\setlength{\tabcolsep}{1mm}
	\small
	\label{tab:results_inpaintguardbench}
	\begin{tabular}{lccccc}
		\toprule
		\multirow{2}{*}{{Methods}} & 
		\multicolumn{5}{c}{{InpaintGuardBench}} \\
		\cmidrule(l){2-6}
		& {PSNR$\downarrow$} & {SSIM$\downarrow$} & {LPIPS$\uparrow$} & {IR$\downarrow$} & {ARC$\downarrow$} \\
		\midrule
		PhotoGuard 
		& 16.385 & 0.610 & 0.397 & 0.147 & 0.685 \\
		AdvDM 
		& 16.354 & 0.604 & 0.398 & 0.102 & 0.677 \\
		MFA 
		& 16.762 & 0.623 & 0.380 & 0.172 & 0.749 \\
		Mist 
		& 15.185 & 0.529 & 0.482 & -0.020 & 0.659 \\
		DiffusionGuard 
		& 14.500 & {0.485} & {0.550} & \textbf{-0.251} & 0.594 \\
		DDD 
		& {14.305} & 0.506 & 0.503 & 0.010 & 0.567 \\
		
		Anti-Inpainting 
		& \textbf{13.364} & \textbf{0.458} & \textbf{0.565} & {-0.131} & \textbf{0.532} \\
		\bottomrule
	\end{tabular}
	\caption{The protective performance of comparison methods and Anti-Inpainting against the different version of diffusion-based inpainters on InpaintGuardBench.}
	\label{table3}
\end{table}

We assessed the transferability of adversarial examples generated by our proposed approach from Runway's v1.5 model to StableAI's v2.0 model. It is noteworthy that while these two models share an identical architecture, they are trained under different protocols. As shown in Table \ref{table3}, our approach demonstrates state-of-the-art performance on the InpaintGuardBench dataset, surpassing all comparison methods. In our approach, we diversify the training conditions by incorporating initial latent state resampling and mask augmentation. The experimental results indicate that this strategy not only enhances the effectiveness of adversarial examples across various conditions but also mitigates the risk of overfitting to specific model weights.
\begin{figure}[!t]
	\centering
	\includegraphics[width=\linewidth]{ 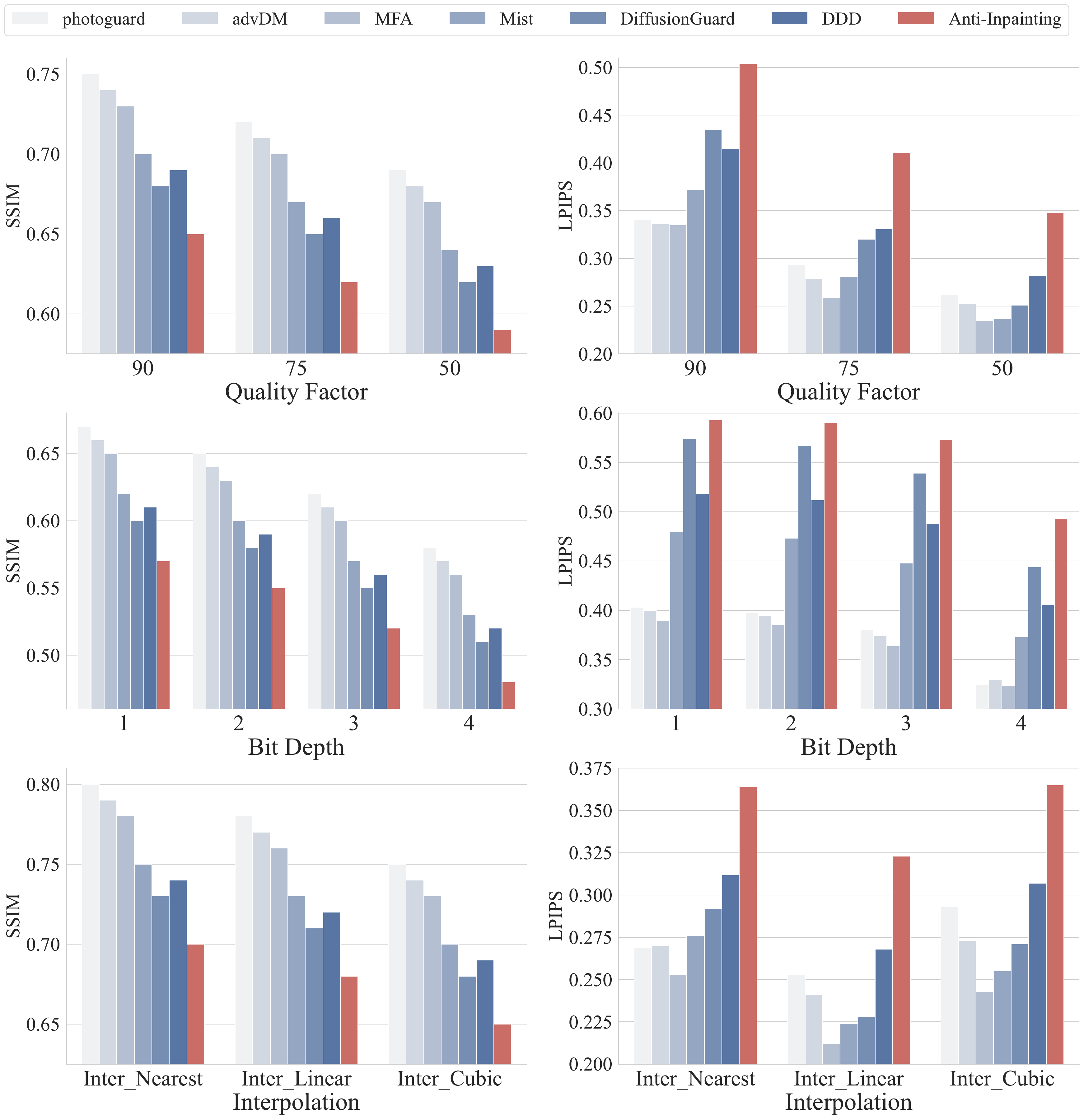}
	\caption{The protective performance of baseline models and Anti-Inpainting through various purification methods.}
	\label{fig6}
\end{figure}
\subsection{Robustness Study}
We conduct the robustness experiments of Anti-Inpainting and other methods against JPEG compression, resizing, and bit depth reduction.
Traditional attacks, targeting only high-frequency features in the U-Net's downsampling blocks, are vulnerable to such purification. In contrast, our method perturbs features across all U-Net levels—downsampling, middle, and upsampling. This ensures that when lossy operations remove high-frequency details, the crucial mid-to-high-level semantic perturbations persist \cite{jeong2025faceshielddefendingfacialimage}. Because diffusion models' image understanding relies on the full feature hierarchy, our comprehensive attack proves more robust. 
The experimental results in Figure \ref{fig6} corroborate our approach, demonstrating consistently superior performance across all three robustness scenarios.
\section{Conclusion}
This paper introduces Anti-Inpainting, a novel proactive defense approach against malicious diffusion-based inpainting. Our approach integrates three key innovations to effectively protect images under unknown operational conditions. Firstly, a multi-level deep feature extractor is utilized to enhance protective efficacy. Secondly, multi-scale semantic-preserving data augmentation is incorporated to improve the transferability of adversarial perturbations across diverse guidance conditions. Finally, a selection-based distribution deviation optimization strategy is developed to dynamically adjust the adversarial noise, mitigating ineffective updates. Extensive experiments demonstrate that Anti-Inpainting is a powerful proactive defense against malicious inpainting manipulations.
\bibliography{AAAI2026}
\end{document}